\documentclass[letterpaper, 10 pt, journal, twoside]{IEEEtran}

\IEEEoverridecommandlockouts

\usepackage{cite}
\usepackage[pdftex]{graphicx}
\graphicspath{{figures/}}
\usepackage{amsmath}
\usepackage{amssymb}
\usepackage{array}
\usepackage{stfloats}
\fnbelowfloat
\usepackage{url}
\usepackage{color}
\usepackage{multirow}
\usepackage{booktabs}
\usepackage{float}

\newcommand{\sourcelink}{%
  \pdfstartlink attr{/Border[0 0 0]} user{/Subtype/Link/A<</S/URI/URI(https://github.com/BavanthaU/m2h_mx.git)>>}%
  {\tt\scriptsize https://github.com/BavanthaU/m2h\_mx}%
  \pdfendlink%
}

\begin{document}

\title{M2H-MX: Multi-Task Semantic and Geometric Perception for Real-Time Monocular 3D Scene Graph Construction}

\author{U.V.B.L. Udugama$^{1}$, George Vosselman$^{1}$, and Francesco Nex$^{1}$%
\thanks{$^{1}$All authors are with the Department of Earth Observation Science,
University of Twente, Enschede, 7522 NH, The Netherlands.
{\tt\small \{b.udugama, george.vosselman, f.nex\}@utwente.nl}}%
\thanks{M2H-MX available at \sourcelink.}%
}

\maketitle

\begin{abstract}
Monocular cameras are attractive for robotic perception because they are lightweight,
low-cost, and easy to deploy. However, real-time spatial understanding from a single RGB
stream remains difficult because metric geometry, semantic consistency, inference latency,
and mapping stability are tightly coupled. This paper presents M2H-MX, a deployment-oriented multi-task front
end that combines a frozen foundation-model encoder with lightweight adaptation,
register-gated global context, and controlled cross-task refinement. Its predictions are
converted through a compact perception-to-mapping interface into inputs for a fixed mapping
pipeline, enabling metric--semantic mapping and downstream 3D scene graph construction.
On NYUDv2, M2H-MX-L achieves the
best performance among the compared multi-task baselines, improving semantic mIoU by
4.06 points and reducing depth RMSE by 9.4\%. Runtime evaluation on an NVIDIA RTX 4080 Super
shows that the full asynchronous perception-to-mapping loop sustains $15$--$20$~Hz. On
ScanNet, M2H-MX reduces average trajectory error by 60.7\% compared with a strong
monocular SLAM baseline while producing cleaner metric--semantic maps. These results show
that dense multi-task perception can provide measurable system-level gains for monocular
robotic mapping.
\end{abstract}


\IEEEpeerreviewmaketitle

\section{Introduction}
\IEEEPARstart{M}{onocular} cameras are widely used in robotics because they are compact,
inexpensive, and simple to integrate. Yet, reliable spatial understanding from a single
RGB stream remains challenging. A robot must recover metric geometry, semantic labels, and
stable map updates from ambiguous monocular observations, while still operating within the
latency budget of real-time mapping and planning. As a result, many spatial mapping
systems continue to rely on RGB-D or LiDAR sensors, or on dense perception models that are
too expensive for practical monocular deployment.

This work examines how much a stronger dense perception front end can improve a monocular
mapping system when the remaining pipeline is kept fixed. We use a Mono-Hydra-based
pipeline~\cite{udugama2023monohydra}, building on metric--semantic mapping systems such
as Kimera~\cite{rosinol2020kimera} and Hydra~\cite{hughes2022hydra}. The state estimation,
mapping, optimization, and scene graph construction modules are kept unchanged; M2H-MX
replaces only the dense monocular perception front end by providing predicted metric depth
and semantic labels from RGB input.

Using learned monocular perception inside this loop imposes constraints that are often
hidden in standalone dense prediction benchmarks. The predictions must be accurate, fast
enough to avoid blocking the mapping backend, stable enough to support tracking and map
fusion, and compatible with the interface expected by the SLAM system. Recent multi-task
dense prediction methods improve monocular depth and semantic segmentation by sharing
complementary cues~\cite{vandenhende2020mti,ye2022inverted,lin2024mtmamba},
while metric--semantic maps and scene graphs provide structured representations for robot
reasoning~\cite{rosinol2020kimera,hughes2022hydra}. However, how modern dense prediction
models should be designed and evaluated for real-time monocular mapping remains
underexplored.

\begin{figure}[!t]
\centering
\makebox[\linewidth][c]{\includegraphics[width=1.05\linewidth]{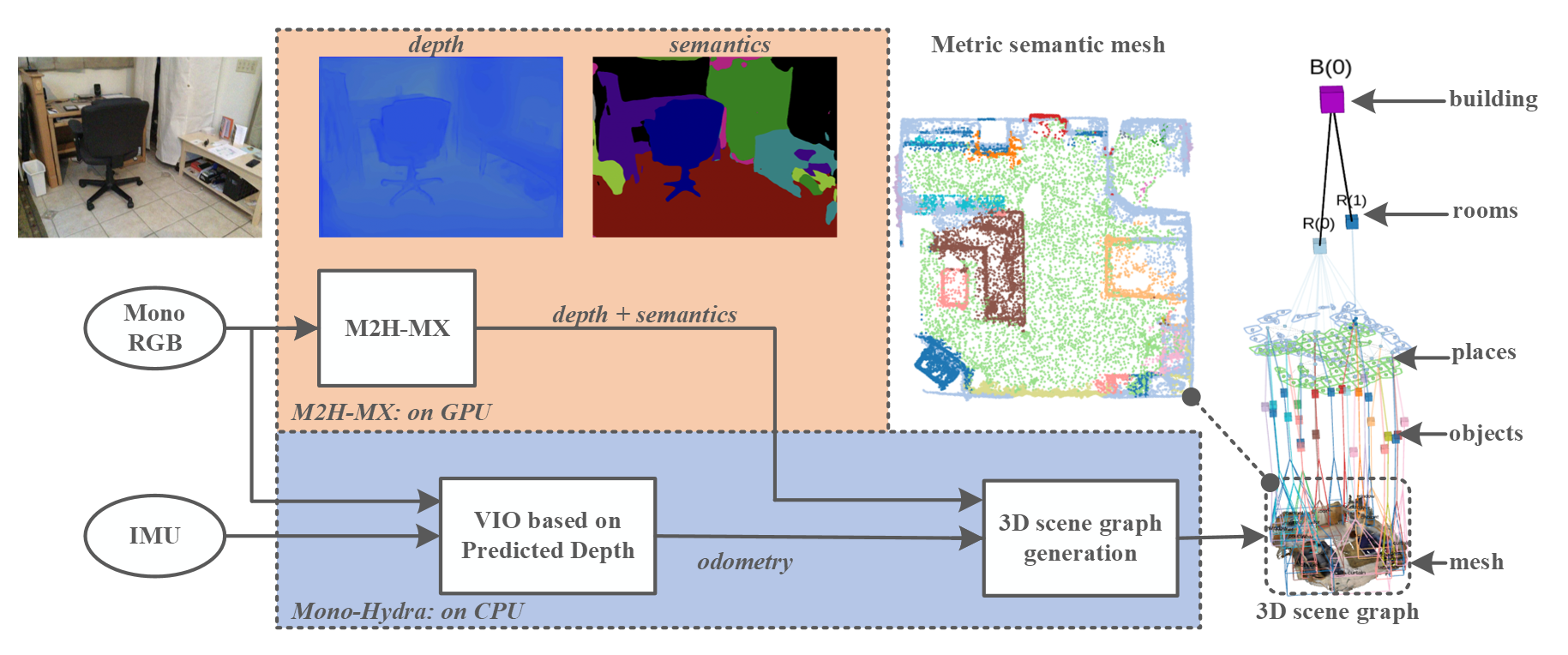}}
\caption{System overview of M2H-MX as a GPU-based perception front end to the fixed
Mono-Hydra~\cite{udugama2023monohydra} monocular SLAM pipeline. From monocular RGB input,
M2H-MX predicts dense depth and semantic labels that are consumed by CPU-based RGB-D
inertial odometry and mapping. The shown pipeline example uses ScanNet scene0000\_00.}
\label{fig:system_overview}
\end{figure}

\begin{figure*}[!t]
\centering
\includegraphics[width=\textwidth]{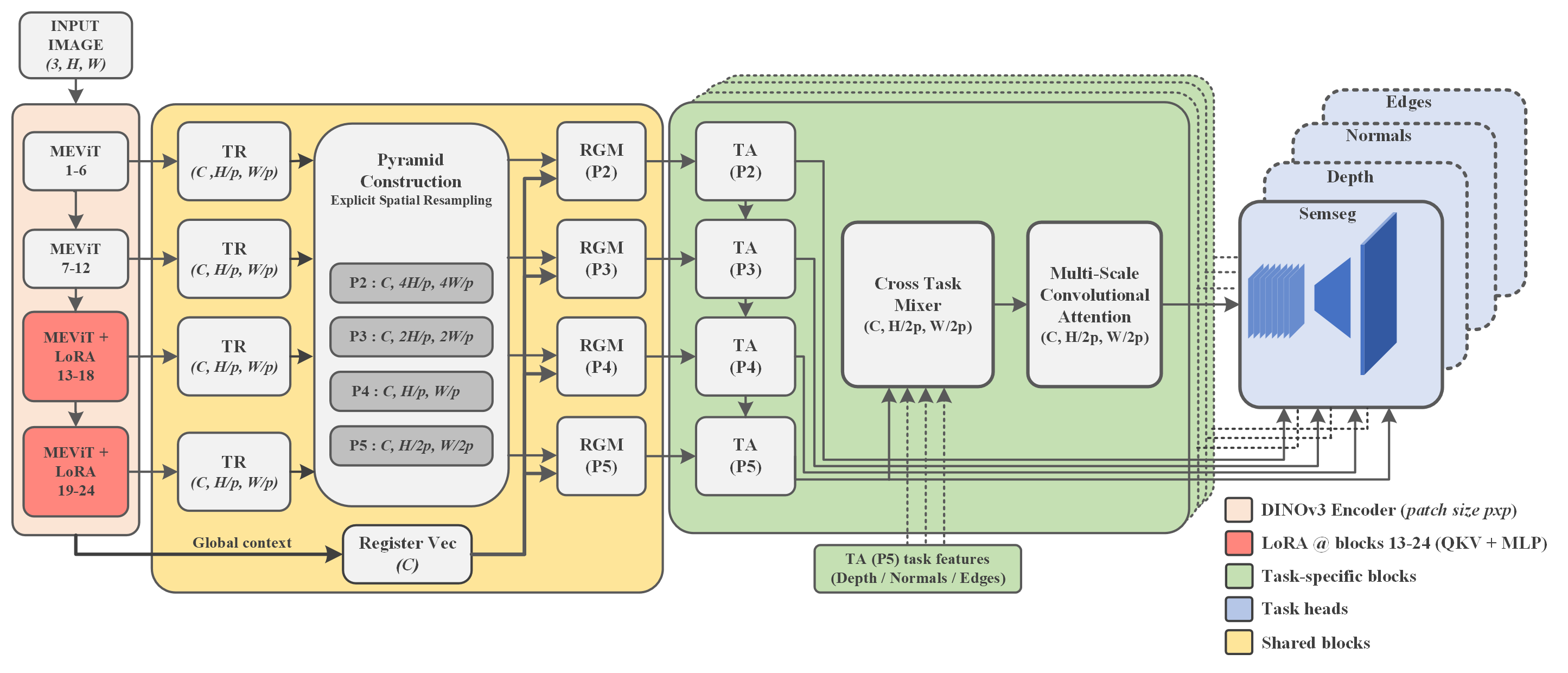}
\caption{Overview of M2H-MX. A monocular RGB image is processed by a DINOv3 encoder with
Memory-Efficient Vision Transformer (MEViT) blocks and LoRA adaptation in the final
transformer blocks. Token reassembly and pyramid construction form multi-scale features.
Register-Gated Mamba (RGM) blocks inject global context from register tokens, while Task
Adaptors (TA), Cross-Task Mixing (CTM), and Multi-Scale Convolutional Attention (MSCA)
produce task-specific features for dense depth, semantic, normal, and edge prediction.}
\label{fig:m2h_hmx_overview}
\end{figure*}

This paper addresses this gap with M2H-MX, a dense monocular perception module designed
for real-time metric--semantic mapping and spatial understanding. Its design is guided by
deployment constraints: preserving strong foundation-model features,
controlling cross-task interference, and maintaining stable runtime behavior.
The main contributions are:
\begin{itemize}
  \item We introduce \textbf{M2H-MX}, a deployment-oriented monocular spatial perception
  framework that predicts metric depth and semantic labels from RGB images while balancing
  accuracy, cross-task consistency, and real-time operation.

  \item We propose a \textbf{compact perception-to-mapping interface} that uses predicted
  depth and semantic cues to support visual-inertial odometry, metric--semantic mapping,
  and downstream \textbf{3D scene graph} generation from monocular RGB input.

  \item We evaluate the resulting system against state-of-the-art monocular SLAM and
  reconstruction methods on ScanNet, addressing the lack of directly comparable monocular
  3D scene graph baselines and demonstrating the downstream value of the proposed
  perception-to-mapping design.
\end{itemize}

\section{Related Work}
Dense multi-task learning has become a common approach for monocular scene understanding,
as jointly predicting geometry and semantics allows complementary task cues to be shared
across representations. Early methods such as MTI-Net~\cite{vandenhende2020mti} introduced structured feature exchange between depth and
semantic segmentation, while transformer-based models such as
InvPT~\cite{ye2022inverted} demonstrated the benefits of global context modeling for dense
prediction. More recent work has explored efficiency-oriented designs that balance accuracy
and computational cost, including state-space and sequence-based decoders such as
MTMamba~\cite{lin2024mtmamba}. Within this line of research, M2H~\cite{udugama2025m2h}
showed that controlled cross-task interaction can improve monocular depth and semantic
prediction while maintaining real-time performance. However, most existing multi-task
models are evaluated primarily in isolation and are not explicitly designed for integration
into a real-time monocular mapping pipeline.

In parallel, metric-semantic mapping systems combine geometric and semantic information to
support higher-level spatial reasoning. Frameworks such as
Kimera~\cite{rosinol2020kimera} and Hydra~\cite{hughes2022hydra} produce structured spatial
representations, including scene graphs, but typically assume depth sensing. Recent
extensions to monocular input, such as Mono-Hydra~\cite{udugama2023monohydra}, replace depth
sensors with learned perception modules, making overall system performance strongly
dependent on the quality, stability, and runtime behavior of dense monocular perception.
M2H-MX differs by designing the dense predictor around the latency, stability, and
interface constraints of a fixed monocular mapping pipeline.

\section{Methodology}
\subsection{M2H-MX Network}
M2H-MX maps a monocular RGB frame to dense geometric and semantic predictions used by the
downstream metric--semantic mapping pipeline. Its design follows four constraints: preserving strong
foundation-model features, limiting trainable parameters, controlling cross-task
interference, and maintaining real-time inference. Given an RGB image
$I_t\in\mathbb{R}^{3\times H\times W}$, the network predicts
$\{\hat{Y}_t^q\}_{q\in\mathcal{K}}=\mathrm{M2H\text{-}MX}(I_t)$, where
$\mathcal{K}=\{\mathrm{depth},\mathrm{sem},\mathrm{norm},\mathrm{edge}\}$. Depth and
semantics are consumed by the mapping pipeline, while normals and edges provide auxiliary
structural supervision.

For compact notation, $\mathrm{Conv}$ denotes convolution, $\mathrm{Up}$
and $\mathrm{Pool}$ denote spatial upsampling and downsampling,
$\mathrm{Reshape}$ denotes map-token conversion, $\sigma$ denotes the
sigmoid function, $\odot$ denotes elementwise multiplication, and $*$
denotes convolution in prediction heads.
As shown in Fig.~\ref{fig:m2h_hmx_overview}, the architecture has four stages: backbone
adaptation and feature pyramid construction, register-gated multi-scale decoding,
task-specific cross-task refinement, and dense prediction heads. The resulting depth and
semantic outputs are passed to the fixed VIO and mapping backend, where pose estimation
and map fusion provide temporal alignment.

\subsubsection{Backbone Adaptation and Shared Feature Pyramid}
The first stage extracts dense features from a frozen DINOv3 encoder~\cite{simeoni2025dinov3}.
In Fig.~\ref{fig:m2h_hmx_overview}, MEViT denotes the Memory-Efficient Vision Transformer
blocks used inside the DINOv3 encoder. For input $I_t$, selected encoder layers
$\mathcal{L}$ produce hidden states $\{H^\ell\}_{\ell\in\mathcal{L}}$. Each $H^\ell$
contains patch tokens $H^\ell_{\mathrm{patch}}\in\mathbb{R}^{N\times D}$, where
$N=(H/p)(W/p)$ is the number of image patches, $p$ is the patch size, and $D$ is the
encoder embedding dimension.

To adapt the frozen encoder with few trainable parameters, LoRA~\cite{hu2021lora} is
applied to the query-key-value (QKV) and multilayer perceptron (MLP) projections of the
final 12 transformer blocks. For a frozen projection $W_0$, LoRA uses
\[
W_{\mathrm{eff}} = W_0 + \lambda BA,
\]
where $A$ and $B$ are trainable low-rank matrices and $\lambda$ is a scaling factor. Thus,
the pretrained backbone remains fixed while the final transformer blocks are adapted for
dense prediction.

Patch tokens are converted into spatial maps using token reassembly (TR) and projected to
a common channel dimension:
\[
\begin{aligned}
F^\ell &= \mathrm{TR}(H^\ell_{\mathrm{patch}}),\\
\widetilde{F}^{\ell} &= \mathrm{Conv}_{1\times1}^{\ell}(F^\ell).
\end{aligned}
\]
The selected reassembled features are fused into
$F_{\mathrm{base}}=\Phi(\{\widetilde{F}^{\ell}\}_{\ell\in\mathcal{L}})$, where
$\Phi(\cdot)$ denotes convolution, normalization, activation, and feature aggregation.
The pyramid construction block then forms the shared decoder pyramid:
\[
\begin{aligned}
p_4 &= \psi_4(F_{\mathrm{base}}),
& p_5 &= \psi_5(\mathrm{Pool}(p_4)),\\
p_3 &= \psi_3(\mathrm{Up}(p_4)),
& p_2 &= \psi_2(\mathrm{Up}(p_3)).
\end{aligned}
\]
Here, $\psi_k$ is a scale-specific projection, while $\mathrm{Pool}$ and $\mathrm{Up}$
denote spatial downsampling and bilinear upsampling. This produces the shared pyramid
$\{p_2,p_3,p_4,p_5\}$.

The final encoder layer also provides $R$ register tokens
$\{h^{\mathrm{last}}_{\mathrm{reg},j}\}_{j=1}^{R}$, shown as the
``Register Vec'' path in Fig.~\ref{fig:m2h_hmx_overview}. These tokens
are pooled into a compact global context vector
\[
r = W_r
\left(
\frac{1}{R}\sum_{j=1}^{R} h^{\mathrm{last}}_{\mathrm{reg},j}
\right),
\]
which conditions all decoder scales with scene-level context.

We use DINOv3 because its Gram anchoring improves dense feature stability during
large-scale self-supervised training, which is important for the spatially coherent patch
features required by M2H-MX~\cite{simeoni2025dinov3}.

\subsubsection{Register-Gated Multi-Scale Decoder}
\begin{figure}[!t]
\centering
\includegraphics[width=0.85\linewidth]{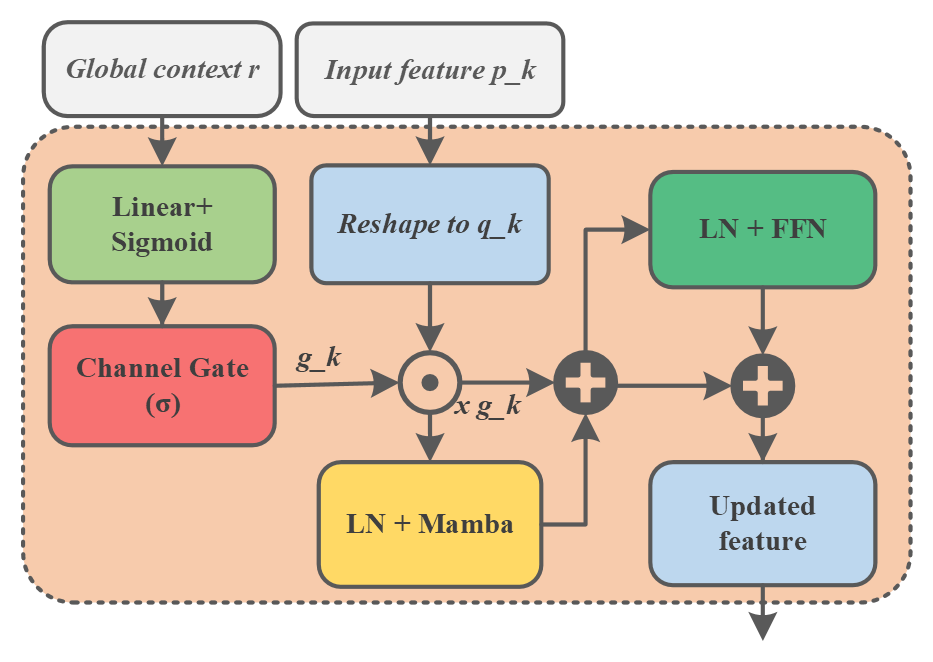}
\caption{Register-Gated Mamba (RGM) block used at each decoder scale. The global context
vector $r$ generates a channel gate $g_k$ through a Linear+Sigmoid projection. The input
feature is reshaped into $q_k$, modulated by $g_k$, and refined by LN+Mamba and LN+FFN
residual branches to produce the updated feature.}
\label{fig:rgm_block}
\end{figure}

The second stage propagates information across pyramid scales using Register-Gated Mamba
(RGM) blocks. At scale $k\in\{5,4,3,2\}$, the pyramid feature is fused with the upsampled
coarser decoder state and reshaped into a token sequence:
\[
\begin{aligned}
x_k &= p_k+\mathrm{Up}(s_{k+1};p_k),\\
q_k &= \mathrm{Reshape}(x_k).
\end{aligned}
\]
Here, $s_{k+1}$ is the decoder output from the next coarser pyramid level, and
$\mathrm{Up}(s_{k+1};p_k)$ upsamples it to the spatial resolution of $p_k$ before fusion.
At the coarsest level, we set $s_6=0$.
The global register vector generates a channel gate,
\[
\begin{aligned}
g_k &= \sigma(\mathcal{A}_k(r)),\\
q_k^g &= q_k\odot g_k,
\end{aligned}
\]
where $\mathcal{A}_k$ is a scale-specific linear projection and $g_k$ is broadcast across
spatial tokens. The gated sequence is then refined by a Mamba block~\cite{gu2023mamba}
and a feed-forward network:
\[
s_k = \mathrm{Reshape}^{-1}\!\left(\mathrm{RGM}_k(q_k,q_k^g)\right).
\]
Here, $s_k\in\mathbb{R}^{C\times H_k\times W_k}$ is the updated shared decoder feature
at scale $k$, and the set $\{s_2,s_3,s_4,s_5\}$ is passed to the task-adaptor blocks in
the next stage. The notation $\mathrm{RGM}_k(q_k,q_k^g)$ summarizes the LN+Mamba and
LN+FFN residual branches in Fig.~\ref{fig:rgm_block}; $q_k^g$
provides global conditioning, while $q_k$ is preserved through residual
updates. LN denotes layer normalization and FFN denotes a feed-forward
network.

\subsubsection{Task-Specific Cross-Task Refinement}
The third stage adapts the shared decoder features $\{s_k\}_{k=2}^{5}$ into task-specific
representations. Task adaptors (TA) convert these shared multi-scale features into
branch-specific features $h_\tau$ for each target task $\tau$ before cross-task
refinement. Cross-Task Mixing (CTM) then injects selected context features into a target branch:
\[
\begin{aligned}
z_j &= \Pi_j(h_j)\odot\big(1+\sigma(G_j(h_j))\big),\\
u_\tau &= \mathrm{Conv}_{1\times1}\big([h_\tau,\{z_j\}_{j\in\mathcal{C}_\tau}]\big).
\end{aligned}
\]
Here, $h_\tau$ is the target-task feature, $h_j$ is a context-task feature,
$\mathcal{C}_\tau$ is the selected context set, $\Pi_j$ aligns channels, and
$G_j$ predicts a gate. This task-directed design allows each branch to use complementary cues only where
useful, instead of forcing all tasks to exchange information symmetrically.

The mixed feature $u_\tau$ is refined by Multi-Scale Convolutional Attention (MSCA), which
uses depthwise convolutions with multiple receptive fields to produce a spatial attention
map $A_{\mathrm{spatial}}^\tau$. The refined feature is
\[
\tilde{h}_\tau = u_\tau + A_{\mathrm{spatial}}^\tau \odot u_\tau.
\]
MSCA emphasizes task-relevant local structures such as boundaries, planar regions, and
thin objects. CTM followed by MSCA is applied to semantic, normal, and edge branches. The
depth branch uses a dedicated bin-based head for adaptive metric calibration and residual
refinement, while its feature can still serve as CTM context for the other branches.

\subsubsection{Task Heads}
The fourth stage converts refined task features into dense predictions. Depth is predicted
with an adaptive bin-based head. Given the refined depth feature $\tilde{h}_d$, global average
pooling predicts adaptive bin widths, while a per-pixel classifier predicts bin probabilities:
\[
w = \mathrm{softmax}(W_w * \mathrm{GAP}(\tilde{h}_d)), \qquad
p_b = \mathrm{softmax}(W_b * \tilde{h}_d).
\]
The bin centers $c_i$ are obtained by cumulatively summing the predicted
bin widths $w$ over the valid depth range. The final depth is
\[
\hat{D} =
\sum_{i=1}^{N_b} p_{b,i}c_i + W_o * \tilde{h}_d.
\]
Here, $\mathrm{GAP}$ is global average pooling, $N_b$ is the number of
adaptive depth bins, $p_{b,i}$ is the probability of bin $i$ at each pixel,
and $W_o * \tilde{h}_d$ is the residual depth correction.

The semantic branch uses a lightweight convolutional classifier:
\[
\hat{S}=\mathrm{Conv}_{1\times1}\!\left(
\delta(\mathrm{Conv}_{3\times3}(\tilde{h}_s))\right),
\]
where $\delta$ denotes the nonlinear activation. The normal and edge branches use
lightweight dense heads:
\[
\hat{N}=\mathrm{NormHead}(\tilde{h}_n), \qquad
\hat{E}=\sigma(\mathrm{EdgeHead}(\tilde{h}_e)).
\]

\subsubsection{Loss Functions and Uncertainty-Based Balancing}
Each task $q\in\mathcal{K}$ uses a main loss and auxiliary losses from decoder scales:
\[
\mathcal{L}_q =
\mathcal{L}^{\mathrm{main}}_q
+
\sum_{k=2}^{5}\alpha_{k,q}\mathcal{L}^{\mathrm{aux}}_{k,q}.
\]
The loss type follows the output: pixel-wise classification for semantics, dense
regression for depth, angular or cosine loss for normals, and binary loss for edges.
Auxiliary supervision encourages intermediate decoder features to remain task-relevant.

We also use consistency terms between geometrically related outputs:
\[
\mathcal{L}_{\mathrm{cons}}
=
\lambda_{dn}\mathcal{L}_{dn}(\hat{D},\hat{N})
+
\lambda_{se}\|\sigma(\hat{E})-\phi(\hat{S})\|_1 .
\]
The first term encourages compatibility between predicted depth and normals, while the
second aligns predicted edges with a semantic-boundary map $\phi(\hat{S})$ computed from
the predicted segmentation.

The full multi-task objective uses learned uncertainty weighting~\cite{kendall2018multi}:
\[
\mathcal{L}_{\mathrm{total}}
=
\sum_{q\in\mathcal{K}}
\left(
\frac{1}{2s_q^2}\mathcal{L}_q+\log s_q
\right)
+
\mathcal{L}_{\mathrm{cons}} .
\]
Here, $s_q$ is the learned uncertainty of task $q$. This reduces manual
tuning of task weights by assigning lower effective weight to tasks with
higher learned uncertainty, while the logarithmic term prevents unbounded
growth of $s_q$.

\subsection{System Integration and Scope}
Fig.~\ref{fig:system_overview} illustrates how M2H-MX is deployed within the Mono-Hydra
monocular SLAM pipeline~\cite{udugama2023monohydra}.
At runtime, M2H-MX replaces only the perception front end. It predicts dense depth and
semantic labels from monocular RGB input; the predicted depth is combined with RGB to form
RGB-D frames for inertial odometry, while semantic labels are used by the mapping backend
for metric--semantic map construction and downstream scene graph generation.
All state estimation, optimization, mapping, and scene graph construction modules are kept
unchanged; therefore, the system-level differences reported in Sec.~IV
can be attributed to the proposed perception front end and its perception-to-mapping
interface.

\section{Experiments}
\noindent The experimental evaluation addresses three questions:
(i) does M2H-MX improve dense multi-task prediction quality,
(ii) do these improvements translate into measurable gains in a running monocular SLAM
system, and
(iii) which architectural components are responsible for these gains.

\subsection{Datasets and Metrics}
We evaluate dense perception performance on NYUDv2~\cite{silberman2012nyu} and
Cityscapes~\cite{cordts2016cityscapes}, and system-level behavior in a running monocular
SLAM pipeline on ScanNet~\cite{dai2017scannet}.
NYUDv2 and Cityscapes represent standard indoor and outdoor benchmarks for joint semantic
and depth estimation, while ScanNet enables evaluation under realistic deployment
conditions.
Reported metrics include semantic mIoU, depth or disparity RMSE, and Absolute Trajectory
Error (ATE) for SLAM evaluation.

\subsection{Implementation Details and Evaluation Protocol}
Unless otherwise stated, experiments use M2H-MX-L with a DINOv3-ViT-L backbone,
decoder width $C=256$, Mamba state size 32, four register tokens, and 64 depth bins.
M2H-MX-B differs only by using the shallower DINOv3-ViT-B backbone; the decoder, task
heads, and training protocol remain unchanged. For M2H-MX-L, LoRA is applied to the final
12 backbone blocks ($r=16$, $\alpha=32$, dropout 0.05), with all other backbone parameters
frozen. Input resolution follows each dataset protocol.

For NYUDv2, all four heads (depth, semantics, normals, edges) are active.
For Cityscapes evaluation and ScanNet deployment, only depth and semantics are enabled
because these datasets do not provide the full set of normal and edge annotations needed
to train and evaluate the auxiliary heads consistently.
ScanNet experiments use a model trained on the ScanNet25k subset, on which it achieves
76.10 mIoU and 0.2210 depth RMSE.
Evaluation sequences are not used for model selection.
Runtime evaluation is performed on a workstation with an Intel i7-14700K CPU, 32GB DDR5
memory, and an NVIDIA RTX 4080 Super GPU. Perception runs asynchronously on the GPU,
while state estimation and mapping run on the CPU.

\subsection{Dense Perception Benchmarks}
We begin by comparing dense prediction quality on standard benchmarks to verify that
M2H-MX improves per-frame depth and semantic estimates. We then test the same model inside
a running monocular SLAM pipeline and ablate the main design blocks.
Table~\ref{tab:nyudv2_results} summarizes NYUDv2 results compared against representative
multi-task learning baselines.
M2H-MX-L improves both semantic and geometric accuracy, achieving the highest mIoU and the
lowest depth RMSE among all compared methods.

Relative to M2H, the strongest compared baseline on NYUDv2, M2H-MX-L improves semantic
mIoU by $+4.06$ points ($61.54 \rightarrow 65.60$) while reducing depth RMSE by
approximately $9.4\%$ ($0.4196 \rightarrow 0.3800$).
These gains indicate that the proposed register-gated decoding and controlled cross-task
interaction improve both prediction quality and cross-task consistency.

\emph{Comparison scope.} Since several compared methods could not use
DINOv3~\cite{simeoni2025dinov3} at the time they were developed, the
NYUDv2 comparison should be interpreted as a complete M2H-MX configuration comparison
rather than a backbone-controlled decoder comparison. The ablation in Table~\ref{tab:backbone_ablation}
therefore shows how much M2H-MX depends on the encoder choice.
\begin{table}[t]
\centering
\footnotesize
\setlength{\tabcolsep}{4pt}
\caption{NYUDv2 depth and semantics results.}
\label{tab:nyudv2_results}
\begin{tabular}{lcc}
\toprule
Method & Semseg mIoU $\uparrow$ & Depth RMSE $\downarrow$ \\
\midrule
TaskPrompter~\cite{ye2023taskprompter} & 55.30 & 0.5152 \\
MQTransformer~\cite{xu2022mqtransformer} & 54.84 & 0.5325 \\
MTMamba~\cite{lin2024mtmamba} & 55.82 & 0.5066 \\
InvPT-B$|$MTPD-C~\cite{ye2022inverted,shang2024bimtdp} & 54.86 & 0.5150 \\
MLoRE~\cite{yang2024multi} & 55.96 & 0.5076 \\
MTMamba++~\cite{lin2025mtmamba++} & 57.01 & 0.4818 \\
M2H~\cite{udugama2025m2h} & 61.54 & 0.4196 \\
\midrule
M2H-MX-B & 61.80 & 0.4170 \\
\textbf{M2H-MX-L (this work)} & \textbf{65.60} & \textbf{0.3800} \\
\bottomrule
\end{tabular}
\end{table}

Table~\ref{tab:cityscapes_results} reports Cityscapes results.
Compared with the strongest baseline, MTMamba++, M2H-MX-L improves semantic mIoU by
$+3.15$ points ($79.13 \rightarrow 82.28$) while reducing disparity RMSE from $4.63$ to
$3.89$.
This demonstrates that the proposed design generalizes beyond indoor datasets and remains
effective in large-scale outdoor scenes.

\begin{table}[t]
\centering
\footnotesize
\setlength{\tabcolsep}{4pt}
\caption{Cityscapes semantic and disparity results.}
\label{tab:cityscapes_results}
\begin{tabular}{lcc}
\toprule
Method & Semseg mIoU $\uparrow$ & Disparity RMSE $\downarrow$ \\
\midrule
MTI-Net~\cite{vandenhende2020mti} & 59.85 & 5.06 \\
InvPT~\cite{ye2022inverted} & 71.78 & 4.67 \\
TaskPrompter~\cite{ye2023taskprompter} & 72.41 & 5.49 \\
MTMamba~\cite{lin2024mtmamba} & 78.00 & 4.66 \\
MTMamba++~\cite{lin2025mtmamba++} & 79.13 & 4.63 \\
\textbf{M2H-MX-L (this work)} & \textbf{82.28} & \textbf{3.89} \\
\bottomrule
\end{tabular}
\end{table}

\subsection{Real-Time System Evaluation in SLAM}
While dense benchmark performance is necessary, the primary objective of M2H-MX is
stable deployment in a real-time monocular SLAM system.
Table~\ref{tab:deploy_runtime} reports model-level profiling in terms of parameters and
GFLOPs.
Despite the additional task heads and cross-task refinement, the M2H-MX-L configuration
requires $371.76$ GFLOPs with the two active deployment heads, depth and semantics, and
$491.91$ GFLOPs with all four heads. In Mono-Hydra, the four-head deployment sustains
$15$--$20$~Hz in the asynchronous perception-to-mapping loop.

\begin{table}[t]
\centering
\scriptsize
\setlength{\tabcolsep}{4pt}
\caption{Model complexity under different active-head settings. GFLOPs are reported at
the evaluated input resolution.}
\label{tab:deploy_runtime}
\begin{tabular}{lcc}
\toprule
Method & \#P (M) & GFLOPs \\
\midrule
\multicolumn{3}{l}{\scriptsize\textit{Reported baselines}} \\
TaskPrompter~\cite{ye2023taskprompter} & 373.00 & 416 \\
MTMamba++~\cite{lin2025mtmamba++} & 315.00 & 524 \\
M2H~\cite{udugama2025m2h} & 81.00 & 488 \\
\midrule
\multicolumn{3}{l}{\scriptsize\textit{M2H-MX variants (this work)}} \\
M2H-MX-B (4 heads) & 134.26 & 322.67 \\
M2H-MX-L (2 heads) & 332.03 & 371.76 \\
M2H-MX-L (4 heads) & 353.53 & 491.91 \\
\bottomrule
\end{tabular}
\end{table}

To evaluate downstream impact, Table~\ref{tab:scannet_avg_ate} reports average ATE on
selected ScanNet sequences.
All methods in Table~\ref{tab:scannet_avg_ate} are evaluated on the same selected ScanNet
sequences. RGB-D methods are included as reference points, while the primary comparison is
against monocular SLAM and reconstruction methods because M2H-MX uses only monocular RGB
input at the sensor level.
On the selected ScanNet sequences, M2H-MX reduces average ATE from $17.59$~cm for
monocular Go-SLAM to $6.91$~cm.
This result indicates that improved per-frame depth and semantic quality directly
translates into more stable camera tracking and map construction.

\begin{table}[t]
\centering
\footnotesize
\setlength{\tabcolsep}{4pt}
\caption{Average ATE [cm] on selected ScanNet sequences (lower is better).}
\label{tab:scannet_avg_ate}
\begin{tabular}{lc}
\toprule
Method & Avg. ATE [cm] \\
\midrule
\multicolumn{2}{l}{\scriptsize\textit{RGB-D baselines}} \\
iMAP~\cite{Sucar:etal:ICCV2021} & 56.21 \\
NICE-SLAM~\cite{Zhu2022CVPR} & 13.05 \\
DROID-SLAM (VO)~\cite{teed2021droid} & 11.59 \\
DROID-SLAM~\cite{teed2021droid} & 7.15 \\
\underline{Go-SLAM~\cite{zhang2023goslam}} & \underline{7.02} \\
\addlinespace[1pt]
\multicolumn{2}{l}{\scriptsize\textit{Monocular methods}} \\
DROID-SLAM (VO)~\cite{teed2021droid} & 63.61 \\
DROID-SLAM~\cite{teed2021droid} & 52.60 \\
Go-SLAM~\cite{zhang2023goslam} & 17.59 \\
\textbf{Mono-Hydra stack~\cite{udugama2023monohydra} (with M2H-MX)} & \textbf{6.91} \\
\bottomrule
\end{tabular}
\end{table}

\begin{figure}[!t]
\centering
\begin{tabular}{@{}cccc@{}}
\includegraphics[width=0.22\linewidth]{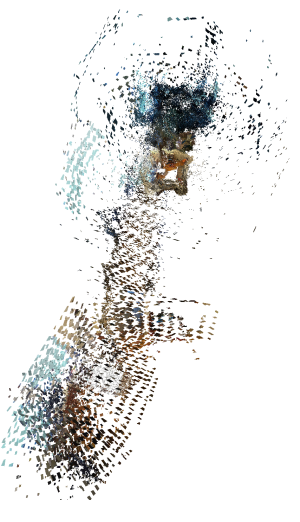} &
\includegraphics[width=0.20\linewidth]{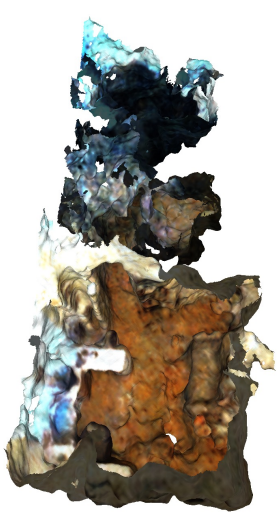} &
\includegraphics[width=0.22\linewidth]{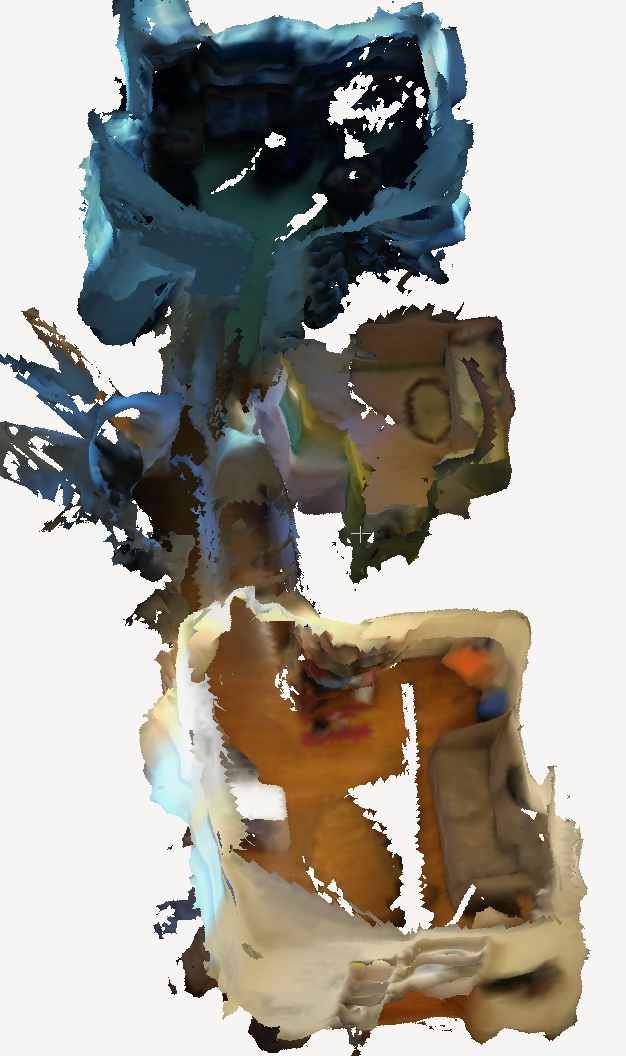} &
\includegraphics[width=0.19\linewidth]{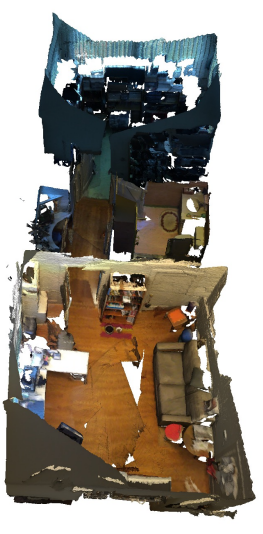} \\
{\scriptsize DROID-SLAM~\cite{teed2021droid}} &
{\scriptsize Go-SLAM~\cite{zhang2023goslam}} &
{\scriptsize M2H-MX} & {\scriptsize GT}
\end{tabular}
\caption{Qualitative monocular mapping comparison on ScanNet scene0054\_00. Compared with
DROID-SLAM and Go-SLAM, integrating M2H-MX produces fewer visible reconstruction artifacts
and more coherent metric--semantic structure in the downstream map.}
\label{fig:deploy_qual}
\end{figure}

Fig.~\ref{fig:deploy_qual} provides a qualitative comparison on ScanNet scene0054\_00,
showing fewer visible reconstruction artifacts and more coherent metric--semantic
structure than the monocular baselines.

\subsection{Ablation Study: Feature Quality vs.\ Decoder Complexity}
Table~\ref{tab:ablation_design} shows that M2H-MX gains arise from a deliberate balance
between strong backbone features and lightweight decoding.
Removing both CTM and MSCA reduces mIoU by 2.07 points and increases depth RMSE, while
CTM-only and MSCA-only variants show limited benefit. Removing RGM or the register feed
also degrades both semantics and depth, confirming the role of register-conditioned
decoding.
\begin{table}[t]
\centering
\scriptsize
\setlength{\tabcolsep}{2.5pt}
\caption{Ablation results on NYUDv2 relative to M2H-MX-L.}
\label{tab:ablation_design}
\label{tab:backbone_ablation}
\begin{tabular}{lcccc}
\toprule
Variant & mIoU $\uparrow$ & RMSE $\downarrow$ & $\Delta$ mIoU & $\Delta$ RMSE \\
\midrule
M2H-MX-L & 65.60 & 0.3800 & -- & -- \\
\midrule
\multicolumn{5}{l}{\scriptsize\textit{Component ablations}} \\
No CTM/MSCA & 63.53 & 0.4619 & -2.07 & +0.0819 \\
CTM only & 63.55 & 0.4705 & -2.05 & +0.0905 \\
MSCA only & 63.53 & 0.4575 & -2.07 & +0.0775 \\
w/o RGM & 64.44 & 0.4346 & -1.16 & +0.0546 \\
w/o reg. feed & 64.22 & 0.4473 & -1.38 & +0.0673 \\
\addlinespace[1pt]
\multicolumn{5}{l}{\scriptsize\textit{Different backbones}} \\
DINOv2-L & 56.79 & 0.5131 & -8.81 & +0.1331 \\
ConvNeXt-L & 38.83 & 0.6940 & -26.77 & +0.3140 \\
\bottomrule
\end{tabular}
\end{table}

\emph{Backbone sensitivity.} Table~\ref{tab:backbone_ablation} also shows that backbone
feature quality matters.
Replacing DINOv3 with ConvNeXt causes a large degradation, indicating that the decoder is
designed for dense transformer features and global register context rather than generic
convolutional features. DINOv2 performs better than ConvNeXt, but still reduces mIoU from
$65.60$ to $56.79$ and increases RMSE from $0.3800$ to $0.5131$. This suggests that DINOv3
provides more suitable spatially coherent features, consistent with its Gram anchoring
strategy~\cite{simeoni2025dinov3}. This supports the intended design of M2H-MX as a decoder
specialized for dense transformer features rather than a backbone-agnostic decoder.

\section{Conclusion}
This paper presented M2H-MX, a deployment-oriented monocular perception model for
real-time metric--semantic mapping and scene graph generation. By combining a frozen
foundation-model backbone, lightweight adaptation, register-gated decoding, and controlled
cross-task interaction, M2H-MX improves dense depth and semantic prediction under runtime
constraints.

On NYUDv2, M2H-MX improves semantic mIoU by 4.06 points and reduces depth RMSE by 9.4\%
over the strongest compared multi-task baseline. In the fixed Mono-Hydra pipeline on ScanNet, it reduces average
trajectory error from $17.59$~cm to $6.91$~cm, a $60.7\%$ improvement, while producing
cleaner metric--semantic maps. These results show that carefully designed multi-task
perception can deliver measurable system-level gains when integrated through a compact
perception-to-mapping interface.

\bibliographystyle{IEEEtran}
\bibliography{IEEEabrv}

\begin{thebibliography}{10}
\providecommand{\url}[1]{#1}
\csname url@samestyle\endcsname
\providecommand{\newblock}{\relax}
\providecommand{\bibinfo}[2]{#2}
\providecommand{\BIBentrySTDinterwordspacing}{\spaceskip=0pt\relax}
\providecommand{\BIBentryALTinterwordstretchfactor}{4}
\providecommand{\BIBentryALTinterwordspacing}{\spaceskip=\fontdimen2\font plus
\BIBentryALTinterwordstretchfactor\fontdimen3\font minus
  \fontdimen4\font\relax}
\providecommand{\BIBforeignlanguage}[2]{{%
\expandafter\ifx\csname l@#1\endcsname\relax
\typeout{** WARNING: IEEEtran.bst: No hyphenation pattern has been}%
\typeout{** loaded for the language `#1'. Using the pattern for}%
\typeout{** the default language instead.}%
\else
\language=\csname l@#1\endcsname
\fi
#2}}
\providecommand{\BIBdecl}{\relax}
\BIBdecl

\bibitem{udugama2023monohydra}
U.~Udugama, G.~Vosselman, and F.~Nex, ``Mono-hydra real-time 3d scene graph
  construction from monocular camera input with imu,'' \emph{ISPRS Annals of
  Photogrammetry, Remote Sensing and Spatial Information Sciences}, vol.
  X-1/W1-2023, pp. 439--445, 2023.

\bibitem{rosinol2020kimera}
A.~Rosinol, M.~Abate, Y.~Chang, and L.~Carlone, ``Kimera: an open-source
  library for real-time metric-semantic localization and mapping,'' in
  \emph{ICRA}.\hskip 1em plus 0.5em minus 0.4em\relax IEEE, 2020, pp.
  1689--1696.

\bibitem{hughes2022hydra}
N.~Hughes, Y.~Chang, and L.~Carlone, ``Hydra: A real-time spatial perception
  system for 3d scene graph construction and optimization,'' \emph{arXiv
  preprint arXiv:2201.13360}, 2022.

\bibitem{vandenhende2020mti}
S.~Vandenhende, S.~Georgoulis, and L.~Van~Gool, ``Mti-net: Multi-scale task
  interaction networks for multi-task learning,'' in \emph{ECCV}.\hskip 1em
  plus 0.5em minus 0.4em\relax Springer, 2020, pp. 527--543.

\bibitem{ye2022inverted}
H.~Ye and D.~Xu, ``Inverted pyramid multi-task transformer for dense scene
  understanding,'' in \emph{ECCV}.\hskip 1em plus 0.5em minus 0.4em\relax
  Springer, 2022, pp. 514--530.

\bibitem{lin2024mtmamba}
B.~Lin, W.~Jiang, P.~Chen, Y.~Zhang, S.~Liu, and Y.-C. Chen, ``Mtmamba:
  Enhancing multi-task dense scene understanding by mamba-based decoders,'' in
  \emph{ECCV}.\hskip 1em plus 0.5em minus 0.4em\relax Springer, 2024, pp.
  314--330.

\bibitem{udugama2025m2h}
U.~Udugama, G.~Vosselman, and F.~Nex, ``M2h: Multi-task learning with efficient
  window-based cross-task attention for monocular spatial perception,'' in
  \emph{IROS}, 2025, pp. 8067--8072.

\bibitem{simeoni2025dinov3}
O.~Sim{\'e}oni, H.~V. Vo, M.~Seitzer, F.~Baldassarre, M.~Oquab, C.~Jose,
  V.~Khalidov, M.~Szafraniec, S.~Yi, M.~Ramamonjisoa, F.~Massa, D.~Haziza,
  L.~Wehrstedt, J.~Wang, T.~Darcet, T.~Moutakanni, L.~Sentana, C.~Roberts,
  A.~Vedaldi, J.~Tolan, J.~Brandt, C.~Couprie, J.~Mairal, H.~J{\'e}gou,
  P.~Labatut, and P.~Bojanowski, ``{DINOv3},'' 2025.

\bibitem{hu2021lora}
E.~J. Hu, Y.~Shen, P.~Wallis, Z.~Allen-Zhu, Y.~Li, S.~Wang, L.~Wang, and
  W.~Chen, ``Lora: Low-rank adaptation of large language models,'' in
  \emph{International Conference on Learning Representations}, 2022.

\bibitem{gu2023mamba}
A.~Gu and T.~Dao, ``Mamba: Linear-time sequence modeling with selective state
  spaces,'' \emph{arXiv preprint arXiv:2312.00752}, 2023.

\bibitem{kendall2018multi}
A.~Kendall, Y.~Gal, and R.~Cipolla, ``Multi-task learning using uncertainty to
  weigh losses for scene geometry and semantics,'' in \emph{Proceedings of the
  IEEE Conference on Computer Vision and Pattern Recognition}, 2018, pp.
  7482--7491.

\bibitem{silberman2012nyu}
N.~Silberman, D.~Hoiem, P.~Kohli, and R.~Fergus, ``Indoor segmentation and
  support inference from rgbd images,'' in \emph{European Conference on
  Computer Vision}.\hskip 1em plus 0.5em minus 0.4em\relax Springer, 2012, pp.
  746--760.

\bibitem{cordts2016cityscapes}
M.~Cordts, M.~Omran, S.~Ramos, T.~Rehfeld, M.~Enzweiler, R.~Benenson,
  U.~Franke, S.~Roth, and B.~Schiele, ``The cityscapes dataset for semantic
  urban scene understanding,'' in \emph{Proceedings of the IEEE Conference on
  Computer Vision and Pattern Recognition}, 2016, pp. 3213--3223.

\bibitem{dai2017scannet}
A.~Dai, A.~X. Chang, M.~Savva, M.~Halber, T.~Funkhouser, and M.~Nie{\ss}ner,
  ``Scannet: Richly-annotated 3d reconstructions of indoor scenes,'' in
  \emph{Proceedings of the IEEE conference on computer vision and pattern
  recognition}, 2017, pp. 5828--5839.

\bibitem{ye2023taskprompter}
H.~Ye and D.~Xu, ``Taskprompter: Spatial-channel multi-task prompting for dense
  scene understanding,'' in \emph{ICLR}, 2023.

\bibitem{xu2022mqtransformer}
Y.~Xu, X.~Li, H.~Yuan, Y.~Yang, and L.~Zhang, ``Multi-task learning with
  multi-query transformer for dense prediction,'' \emph{IEEE Transactions on
  Circuits and Systems for Video Technology}, vol.~34, no.~2, pp. 1228--1240,
  2024.

\bibitem{shang2024bimtdp}
Y.~Shang, D.~Xu, G.~Liu, R.~R. Kompella, and Y.~Yan, ``Efficient multitask
  dense predictor via binarization,'' in \emph{CVPR}, 2024, pp.
  15\,899--15\,908.

\bibitem{yang2024multi}
Y.~Yang, P.-T. Jiang, Q.~Hou, H.~Zhang, J.~Chen, and B.~Li, ``Multi-task dense
  prediction via mixture of low-rank experts,'' in \emph{CVPR}, 2024, pp.
  27\,927--27\,937.

\bibitem{lin2025mtmamba++}
B.~Lin, W.~Jiang, P.~Chen, S.~Liu, and Y.-C. Chen, ``Mtmamba++: Enhancing
  multi-task dense scene understanding via mamba-based decoders,'' \emph{IEEE
  Transactions on Pattern Analysis and Machine Intelligence}, 2025.

\bibitem{Sucar:etal:ICCV2021}
E.~Sucar, S.~Liu, J.~Ortiz, and A.~Davison, ``{iMAP}: Implicit mapping and
  positioning in real-time,'' in \emph{Proceedings of the International
  Conference on Computer Vision ({ICCV})}, 2021.

\bibitem{Zhu2022CVPR}
Z.~Zhu, S.~Peng, V.~Larsson, W.~Xu, H.~Bao, Z.~Cui, M.~R. Oswald, and
  M.~Pollefeys, ``Nice-slam: Neural implicit scalable encoding for slam,'' in
  \emph{Proceedings of the IEEE/CVF Conference on Computer Vision and Pattern
  Recognition (CVPR)}, 2022.

\bibitem{teed2021droid}
Z.~Teed and J.~Deng, ``{DROID-SLAM: Deep Visual SLAM for Monocular, Stereo, and
  RGB-D Cameras},'' \emph{Advances in neural information processing systems},
  2021.

\bibitem{zhang2023goslam}
Y.~Zhang, F.~Tosi, S.~Mattoccia, and M.~Poggi, ``Go-slam: Global optimization
  for consistent 3d instant reconstruction,'' in \emph{Proceedings of the
  IEEE/CVF International Conference on Computer Vision (ICCV)}, October 2023.

\end{thebibliography}

\end{document}